\documentclass[runningheads]{llncs}

\usepackage{eccv}

\usepackage{eccvabbrv}

\usepackage{graphicx}
\usepackage{booktabs}
\usepackage{multirow}
\usepackage[accsupp]{axessibility}  

\usepackage{hyperref}
\usepackage{orcidlink}

\begin{document}

\title{FDM-MFVT: Few-step Sampling Diffusion Model for Mask-Free Virtual Try-On} 

\titlerunning{FDM-MFVT}

\author{Jiaxin Liu\inst{1}\orcidlink{0009-0000-0624-7783} \and
Xiaoye Liang\inst{1,2}\orcidlink{0009-0004-5284-5504} \and
Lai Jiang\inst{1,3}\orcidlink{0000-0002-4639-8136} \and
Mai Xu\inst{1,3}\orcidlink{0000-0002-0277-3301} \and
Jun Liu\inst{1}\orcidlink{0000-0001-6422-2911}\thanks{Corresponding author}}

\authorrunning{J, ~Liu et al.}

\institute{School of Electronic Information Engineering, Beihang University, Beijing, China \and
Zhongguancun Academy, Beijing, China  \and
State Key Laboratory of Virtual Reality Technology and Systems, Beihang University, Beijing, China\\
\email{liujiaxinkw@buaa.edu.cn}, \email{liujun2019@buaa.edu.cn}
}

\maketitle

\begin{figure*}[]
	\centering
	\includegraphics[width=1\linewidth]{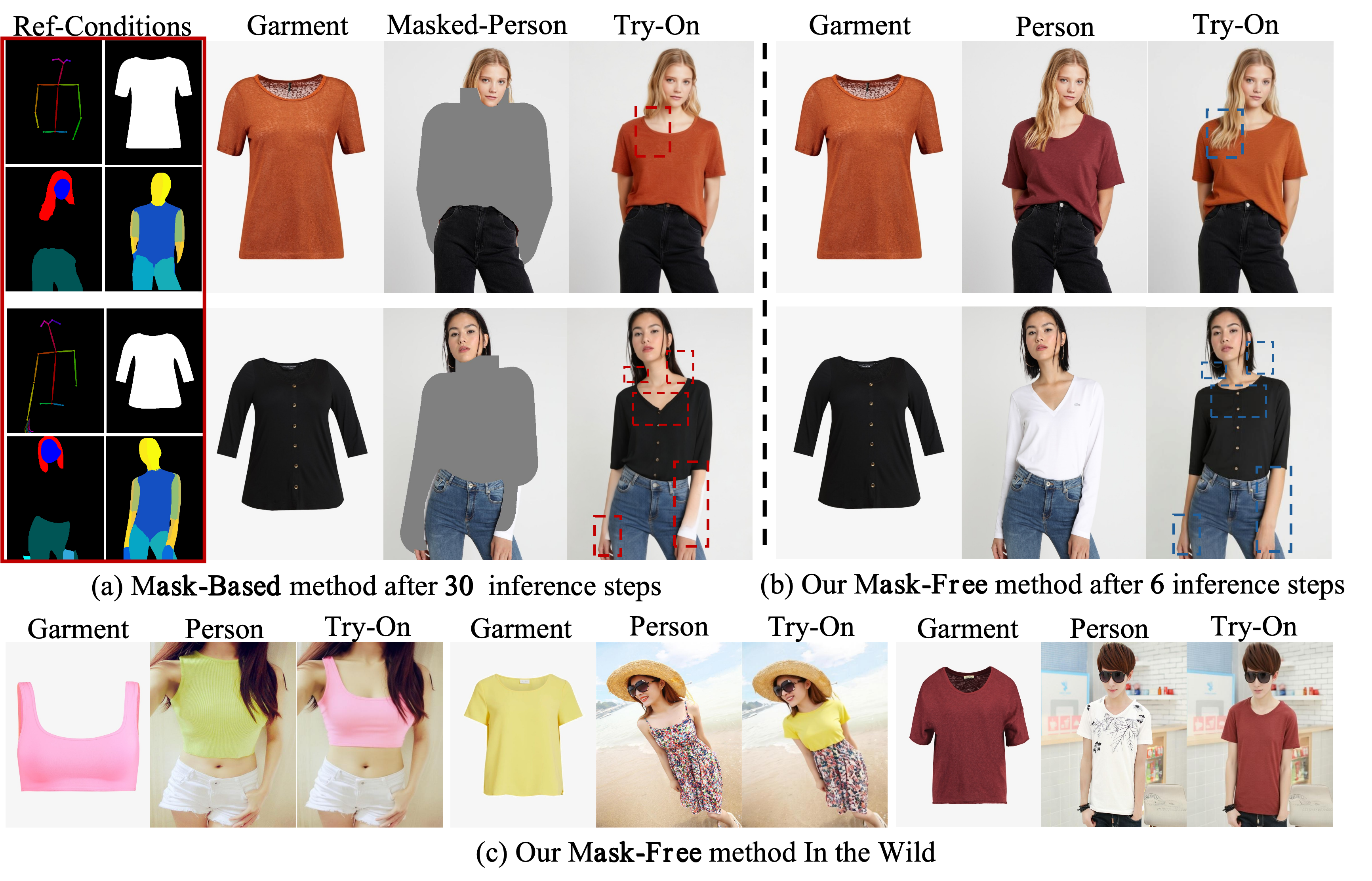}
	\caption{We propose FDM-MFVT, a mask-free few-step virtual try-on framework that achieves high-fidelity virtual try-on with fewer inference steps across multiple scenarios and requires no reference conditions.}
	\label{fig:figure1}
\end{figure*}

\begin{abstract}
  Image-based Virtual Try-On (IVTON) has greatly advanced through diffusion models, yet existing methods require many sampling steps and depend on masks with costly auxiliary networks. In addition, the absence of large-scale mask-free paired datasets further limits the development of mask-free IVTON. We propose FDM-MFVT, a few-step diffusion model for mask-free IVTON, integrating an  Outfit-aware Noise Optimization Module (OANO) and an Instruction-driven Try-on Module (IDT) to enhance efficiency and flexibility.The OANO module initializes the alignment space with noise using the input image and only needs 6 steps to generate a higher-fidelity try-on image compared to 30 steps.The IDT module uses virtual try-on prompts and efficient adaptation to generate high-quality results from garment and person images alone. We further introduce MFVT, a 30,000-pair mask-free IVTON dataset. Experiments show that FDM-MFVT achieves superior quantitative and qualitative results with fewer inference steps than mask-based and mask-free baseline methods.
  \keywords{Diffusion model \and Image synthesis \and Virtual Try-On}
\end{abstract}

\section{Introduction}
\label{sec:intro}
The objective of image-based virtual try-on (IVTON) is to synthesize a realistic image of an individual wearing a target garment, given a person image and a garment image. This task has attracted considerable attention due to its extensive applications in e-commerce, including online retail \cite{hadi2015buy} and virtual fashion design \cite{de2023disentangling,hsiao2018creating,sarkar2023outfittransformer}. To produce photorealistic results, deep learning-based methods \cite{ge2021parser,han2018viton,issenhuth2020not,lee2022high,wang2018toward} typically rely on datasets \cite{choi2021viton,morelli2022dress} comprising aligned tuples of person images, garment images, agnostic-mask, and other conditional inputs.

Recent advances in diffusion models \cite{rombach2022high,berthelot2023tract} have demonstrated exceptional performance in image synthesis across a wide range of tasks. Compared with earlier generative adversarial network (GAN)-based methods \cite{han2018viton,han2019clothflow,he2022style,lee2022high,minar2020cp}, diffusion-based virtual try-on methods \cite{hong2025ita,chong2024catvton,kim2024stableviton,choi2024improving,yang2024texture,xu2025ootdiffusion,liang2026vtedit} achieve substantially higher image fidelity by leveraging the strong generative priors of large-scale diffusion networks. These priors help preserve garment details and improve overall visual quality. However, diffusion models generally require many sampling steps to generate high-fidelity IVTON outputs, which greatly limits their practicality for real-time try-on systems.
In addition to high sampling costs, current diffusion-based virtual try-on systems typically depend on garment masks and multiple auxiliary encoders to process conditional inputs such as clothing masks, DensePose \cite{guler2018densepose}, and human parsing maps. This multi-condition design significantly increases inference complexity and makes the system highly sensitive to mask quality. Furthermore, progress toward mask-free try-on is hindered by the lack of large-scale, high-quality paired datasets without mask annotations.
All in all, existing IVTON methods face three key challenges: 
(1) High computational overhead caused by the many sampling  steps required for high-fidelity diffusion-based synthesis; 
(2) Heavy reliance on masks and auxiliary encoders, leading to complex inference pipelines and high sensitivity to preprocessing errors;
(3) Limited availability of mask-free paired datasets, restricting the development of fully mask-free try-on frameworks.

To address these challenges, we propose FDM-MFVT, a Few-Step Sampling Diffusion Model for Mask-Free Virtual Try-On. We first observe that the random noise distribution in diffusion inference plays a critical role in both synthesis quality and the required number of sampling steps. Certain noise patterns enable high-quality generation with substantially fewer steps, highlighting the importance of effective initialization. Based on this insight, we design an Outfit-aware Noise Optimization Module (OANO) that optimizes the random noise distribution conditioned on the input images. This allows the model to start from a more favorable noise state and produce high-quality results with significantly fewer diffusion steps.
Second, to streamline inference and completely remove the dependence on masks or additional conditional inputs, we introduce the Instruction-driven Try-on Module (IDT). This module incorporates simulated contextual cues and a lightweight adaptation mechanism, enabling the model to infer all necessary information directly from person and garment images. This design maintains high synthesis quality under few-step sampling while substantially reducing architectural complexity.
Finally, to support mask-free IVTON research, we construct MFVT, a large-scale, high-quality mask-free dataset. Extensive experiments show that FDM-MFVT achieves state-of-the-art performance across multiple metrics while significantly improving inference efficiency. Compared with existing mask-free and mask-based VTON methods, our method exhibits superior robustness and real-time capability, as shown in Fig. \ref{fig:figure1}.
In summary, our contributions are threefold:
\begin{itemize}
\item We propose a mask-free IVTON framework that eliminates dependency on auxiliary conditions and operates effectively with only a few inference steps, enabling practical real-time virtual try-on.
\item We design a novel initialization and conditioning strategy that improves diffusion efficiency and garment–person alignment by integrating outfit-aware noise learning with instruction-driven try-on for robust, high-quality synthesis.
\item We construct a new large-scale mask-free dataset (MFVT) that fills a critical gap in training mask-free IVTON systems and provides a solid foundation for future research.
\end{itemize}

\section{Related Work}

\subsection{Image-based Virtual Try-On}
Image-based Virtual Try-On aims to synthesize a realistic image of a person wearing a target garment. Early works such as VITON \cite{han2018viton} introduces a coarse-to-fine GAN framework that transfers multi garments onto human images while preserving pose and garments structure, and VITON-HD \cite{choi2021viton} further enhances realism and resolution through a multi-scale generator. Subsequent methods, including ClothFlow \cite{han2019clothflow}, modeled garment deformation as dense flow fields, while OOTD \cite{xu2025ootdiffusion} provided a large-scale benchmark dataset and methods that rely on human parsing maps and skeleton information. More recently, PICTURE \cite{ning2024picture} enables photorealistic try-on from unconstrained garment inputs, demonstrating strong generalization to real-world clothing. StableVITON \cite{kim2024stableviton}  leveraged stable diffusion model to improve visual fidelity using multiple conditional inputs such as masks and label maps. However, these methods depend on  multi-conditional guidance, which significantly increases inference complexity and limits scalability.
To reduce reliance on explicit conditions, a few studies have explored mask-free virtual try-on. For example, OmniTry \cite{feng2025omnitry} proposed a unified mask-free framework that generalizes virtual try-on to clothing and diverse wearable items. However, current mask-free methods often suffer from blurry garment boundaries, loss of fine texture, and limited robustness across diverse poses. Thus, achieving high-quality mask-free virtual try-on remains an open and challenging research problem.
\subsection{Few-Step Inference in Diffusion}

Diffusion models achieve impressive generative quality but typically require hundreds of denoising steps, which limits their efficiency. To address this, recent works have explored few-step inference strategies, which can be broadly categorized into three directions, e.g. distillation-based \cite{li2023snapfusion,salimans2022progressive,song2023consistency,gu2023boot,berthelot2023tract}, sampler-reformulation-based \cite{song2020denoising} and noise-optimization-based \cite{zhou2025golden} methods. Early acceleration methods focused on distilling multi-step diffusion trajectories into compact student models. Progressive distillation \cite{salimans2022progressive} compresses the sampling process into fewer iterations, while consistency models \cite{song2023consistency} directly learn mappings that enable one-step or few-step generation. These methods demonstrate that knowledge transfer from teacher to student can significantly reduce inference cost while maintaining fidelity.
Another line of work reformulates the reverse process to allow deterministic or non-Markovian sampling. DDIM \cite{song2020denoising} introduced implicit models that reduce the number of steps by redefining the reverse diffusion dynamics, enabling efficient few-step sampling with controllable interpolation. Such mapping-based formulations highlight that the reverse process can be approximated without sacrificing sample diversity.Beyond distillation and mappings, recent studies investigate how noise scheduling and initialization affect efficiency. The Golden Noise method \cite{zhou2025golden} proposes optimized noise distributions that accelerate convergence and improve sample quality in few-step regimes. 
This perspective emphasizes that careful design of noise priors can complement architectural or training-based acceleration.
\begin{figure*}[]
	\centering
	\includegraphics[width=1\linewidth]{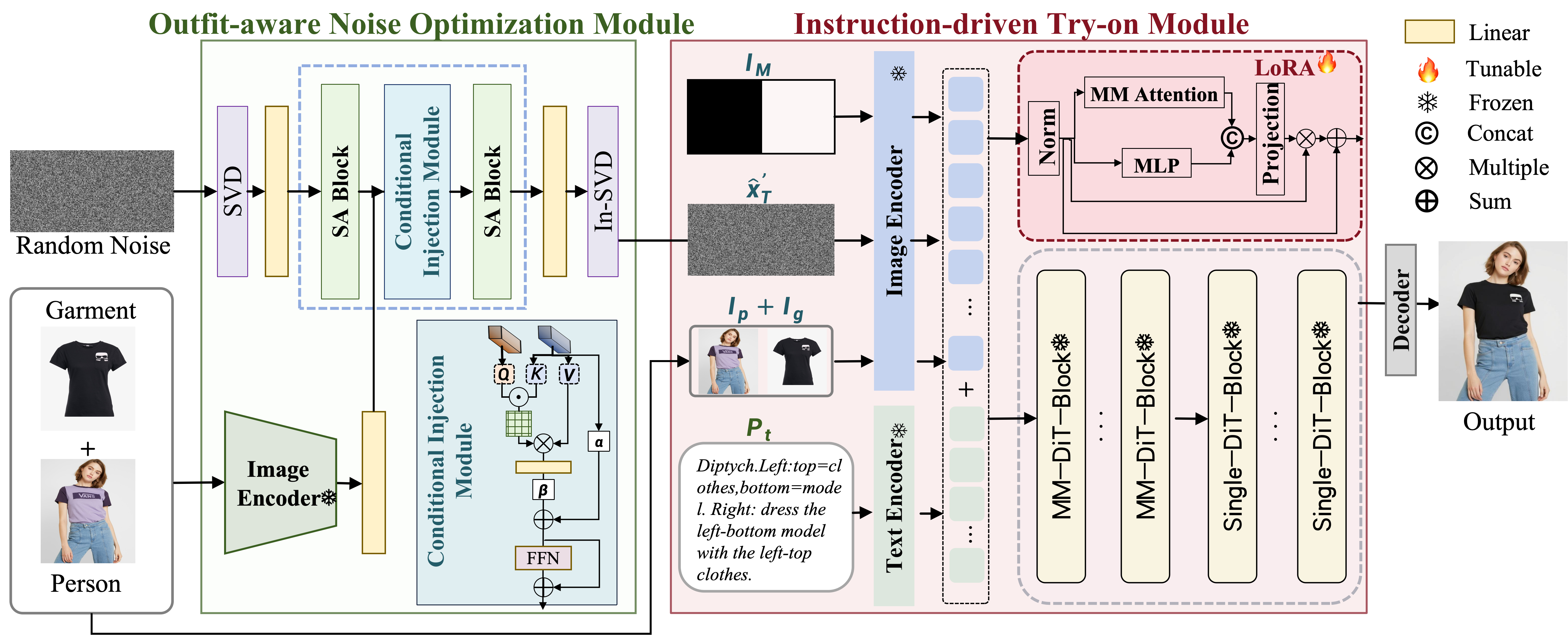}
	\caption{Illustration of the FDM-MFVT framework. The left part shows the Outfit-aware Noise Optimization (OANO) module, which refines noise initialization. The right part shows the Instruction-driven Try-on (IDT) module, which integrates visual and textual features for mask-free synthesis.}
	\label{fig:figure2}
\end{figure*}

\section{Method}
\label{sec:blind}
\subsection{Overview}
In this section, we introduce the FDM-MFVT framework, designed to produce high-quality mask-free virtual try-on results with minimal inference steps.
As illustrated in Fig. \ref{fig:figure2}, our framework consists of an OANO module and an IDT module. 
Within the OANO module, given the garment image $I_g$ and the person image $I_p$, the original noise $x_T$ is refined into optimized noise $x_T^{\prime}$. This optimization leverages outfit-specific features to provide a more informative initialization, thereby improving the fidelity and realism of subsequent synthesis while enabling high-quality outputs with minimal steps. Building on the optimized noise $x_T^{\prime}$, the IDT module integrates both visual and textual instructions to guide the try-on process. Specifically, multi-modal encoders extract features from the garment image $I_g$, person image $I_p$ and mask $I_M$, as well as textual prompts $P_t$, which describes the desired composition. These features are fused through attention-based mechanisms and low-rank adaptation layers, enabling precise alignment between clothing and body structure. The resulting representation is progressively refined through MM-DiT and Single-DiT blocks before being decoded into the final try-on image.

\subsection{OANO Module}
\label{sec:Outfit-aware Noise Optimization Module}
The OANO module is designed to refine noise initialization for mask-free virtual try-on.  
Instead of relying solely on random noise, OANO module constructs noise pairs $(x_T, x_T^{\prime})$ by combining garment and person features with image inversion\cite{avrahami2025stable,cao2023masactrl,meng2021sdedit,wang2024taming}.
These noise pairs provide more informative starting points, capturing structural and semantic features of the input.  
By training on directly generated noise pairs, OANO module produces optimized noise $\widehat{x}_T^{\prime}$ that enhances the realism and consistency of the final synthesis.

\noindent\textbf{Noise Pair Construction}  
Prior studies \cite{qi2024not,meng2023distillation} suggest that not all noises are equally effective, and the choice of initialization strongly influences generation quality. In our framework, noise pairs are directly constructed rather than learned. Specifically, given an input image, we perform inversion to obtain a target noise representation $x_T^{\prime}$ that encodes richer semantic features aligned with the garment and person context. This target noise serves as a more informative initialization compared to purely random noise $x_T$.  
We then filter the generated noise pairs by evaluating reconstruction quality: only those pairs in which $x_T^{\prime}$ yields higher fidelity than $x_T$ are retained. Formally, the construction process is expressed as:
\begin{equation}
\mathcal{D}_{\text{noise}} = \left\{ \left(x_T, x_T^{\prime}\right) \;\middle|\; 
Q(x_T^{\prime}) > Q(x_T) \right\},
\end{equation}
where $Q(\cdot)$ denotes the quality evaluation function. This ensures that the dataset emphasizes informative, outfit-aware noise samples.

\noindent\textbf{Model Architecture}  
Analysis \cite{zhou2025golden} indicates that the singular vectors of $x_T$ and $x_T^{\prime}$ exhibit strong similarity, which can be exploited for efficient noise optimization. Building on this property, the OANO module takes garment and person images $\left(I_g, I_p\right)$ together with random noise $x_T$ as inputs, and outputs optimized noise $\widehat{x}_T^{\prime}$. Specifically, we combine $I_g$ and $I_p$ to obtain $I_{gp}$, encode its features, and perform singular value decomposition (SVD) on $x_T$:
\begin{align}
I_f     &= L\left(E\left(I_{gp}\right)\right), \\
x_T     &= U \times \Sigma \times V^T, \\
x_{T_f} &= L\left(\phi\left(U, \Sigma, V^T\right)\right).
\end{align}
where $E(\cdot)$ denotes the image encoder, $L(\cdot)$ a linear layer, and $\phi(\cdot)$ a ternary function integrating SVD components.  
The Conditional Injection Module then perturbs $x_{T_f}$ using $I_f$ through learnable attention-based operations:
\begin{align}
\boldsymbol{V}_T &= SA(x_{T_f}) \boldsymbol{W}^V, \\
\boldsymbol{K}_T &= SA(x_{T_f}) \boldsymbol{W}^K,\\
\boldsymbol{Q}_R &= I_f \boldsymbol{W}^Q,
\end{align}
\begin{equation}
\boldsymbol{Z}_T = \operatorname{softmax}\left(\frac{\boldsymbol{Q}_R \boldsymbol{K}_T^T}{\sqrt{D_K}}\right) \cdot \boldsymbol{V}_T,
\end{equation}
\begin{equation}
x_{T_f}^{\prime} = \alpha Z_T W^E + \beta x_{T_f} + FFN\left(\alpha Z_T W^E + \beta x_{T_f}\right),
\end{equation}
where $SA(\cdot)$ denotes self-attention, $W^E$ is the output weight matrix, and $\alpha, \beta$ are learnable parameters. Finally, the optimized noise is obtained as:
\begin{equation}
\widehat{x}_T^{\prime} = U \times L\left(SA\left(x_{T_f}^{\prime}\right)\right) \times V^T.
\end{equation}

\noindent\textbf{Training}  
The OANO module is trained on directly generated noise pairs $\mathcal{D}_{\text{noise}}=\{(x_T, x_T^{\prime}, I_g, I_p)\}$, where $x_T$ is random noise and $x_T^{\prime}$ is the target noise obtained via inversion and quality-based filtering. For each training instance, the module predicts $\widehat{x}_T^{\prime}$, supervised to approximate $x_T^{\prime}$ using mean squared error (MSE):
\begin{equation}
\mathcal{L}_{\mathrm{OANO}} = \mathrm{MSE}\left(x_T^{\prime}, \widehat{x}_T^{\prime}\right)
= \frac{1}{D}\left\|x_T^{\prime} - \widehat{x}_T^{\prime}\right\|_2^2,
\end{equation}
where $D$ is the dimensionality of the noise tensor. This objective ensures that $\widehat{x}_T^{\prime}$ closely approximates $x_T^{\prime}$, providing a robust initialization for downstream synthesis.

\subsection{IDT Module}
\label{sec:Instruction-driven Try-on Module}
\noindent\textbf{Model Architecture} 
The IDT module is built upon a transformer-based denoising diffusion framework, leveraging the contextual reasoning ability of large-scale Diffusion Transformers (DiTs) \cite{huang2024context}. Unlike mask-based methods, the IDT module enables mask-free virtual try-on by embedding both visual and textual modalities into a unified contextual prompt. 
Specifically, the garment image $I_g$ , the person image $I_p$, and the mask $I_M$ are first encoded using a variational autoencoder (VAE) encoder:
\begin{equation}
z_i = \mathrm{VAE}\left(\mathrm{Concat}(I_g,I_p)\odot I_M \right),
\end{equation}
where $z_i$ denotes the latent representations of the garment, person, and mask images.
The textual instruction prompt $P_t$ is encoded using a pretrained T5 encoder:
\begin{equation}
z_t = \mathrm{T5}(P_t),
\end{equation}
where $z_t$ represents the textual embedding of the try-on instruction.
All latent tokens are then concatenated to form the contextual input sequence:
\begin{equation}
Z = \mathrm{Concat}(z_i, z_t).
\end{equation}
This unified token sequence $Z$ is subsequently fed into a frozen Inpainting Diffusion Transformer block $D_I$, together with the noisy latent $X$, to guide the try-on process:
\begin{equation}
I_t = D_I(Z, X).
\end{equation}

\noindent\textbf{LoRA Architecture}  
To adapt the frozen DiT blocks for instruction-driven try-on, we employ Low-Rank Adaptation (LoRA). The encoded token sequence $Z$ is first normalized:
\begin{equation}
Z_n = \mathrm{Norm}(Z).
\end{equation}
The normalized tokens are then processed by two parallel branches: multimodal attention (MM-Attn) and a multilayer perceptron (MLP):
\begin{equation}
Z_{attn} = \mathrm{MM\text{-}Attn}(Z_n), \quad Z_{mlp} = \mathrm{MLP}(Z_n).
\end{equation}
The outputs of these branches are concatenated and projected through a linear layer:
\begin{equation}
Z_{proj} = \mathrm{Linear}\left(\mathrm{Concat}(Z_{attn}, Z_{mlp})\right).
\end{equation}
Finally, the projected features are combined with the normalized input via element-wise multiplication and residual addition:
\begin{equation}
Z_{out} = Z_{proj} \odot Z_n + Z_n,
\end{equation}
where $\odot$ denotes element-wise multiplication. The resulting $Z_{out}$ serves as the LoRA-adapted output, enabling fine-tuning of the otherwise frozen DiT blocks.

\noindent\textbf{Training}  
The training of IDT module with LoRA fine-tuning follows the standard denoising diffusion objective. Given a clean target latent $x_0$ and a noisy latent $x_t$ at timestep $t$, the model predicts the added noise $\hat{\varepsilon}_\theta$. The loss function is defined as:
\begin{equation}
\mathcal{L}_{\text{diff}} = \mathbb{E}_{x_0, t, \varepsilon}\left[\left\|\varepsilon - \hat{\varepsilon}_\theta(x_t, t, Z_{out})\right\|^2\right],
\end{equation}
where $\varepsilon$ is the sampled Gaussian noise, and $\hat{\varepsilon}_\theta$ is the noise predicted by the LoRA-adapted DiT block conditioned on $Z_{out}$.
In practice, this objective ensures that the LoRA parameters learn to align garment, person, and mask features with textual instructions, while keeping the majority of the DiT parameters frozen. This design achieves efficient fine-tuning with minimal additional parameters, enabling instruction-driven virtual try-on in a zero-shot setting.

\section{Experiments}
\subsection{Experimental Setting}
\label{sec:ex setting}
\begin{figure}[!t]
	\centering
	\includegraphics[width=1\linewidth]{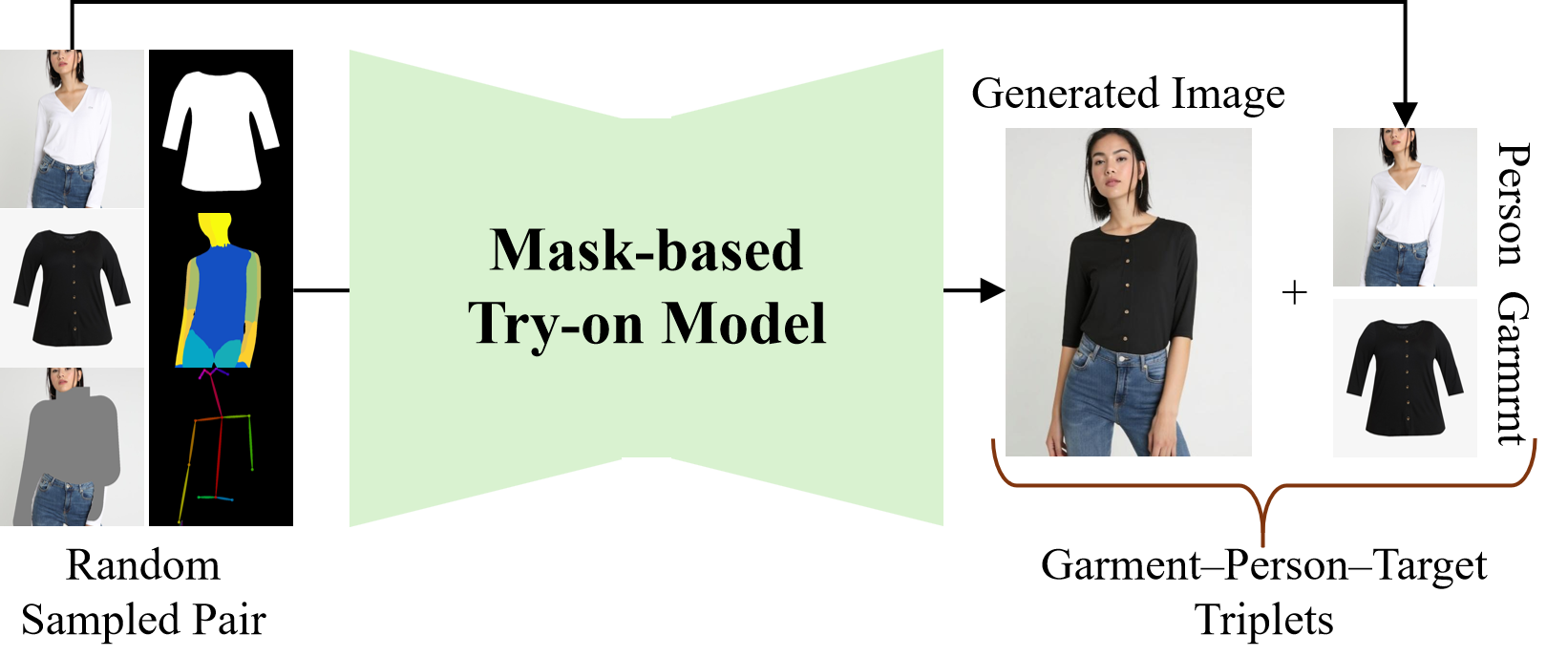}
	\caption{Construction pipeline of the MFVT dataset.}
	\label{fig:data}
\end{figure}

\noindent\textbf{Datasets} 
To facilitate mask-free virtual try-on, we construct a large-scale and high-quality dataset, termed MFVT. The dataset provides garment–person–target triplets without explicit segmentation masks, overcoming the limitations of existing mask-based datasets. As shown in Fig. \ref{fig:data}, MFVT is built by leveraging a pretrained mask-based virtual try-on model to generate person images conditioned on randomly sampled garment–person pairs from existing open-sourced mask-based datasets \cite{choi2021viton,morelli2022dress}. For each sampled pair, the model infers a synthesized person image wearing the target garment.  The generated image, together with the original person and garment images, forms the triplets in MFVT. In total, MFVT contains 30,000 pairs across diverse clothing categories: 10,671 upper-body, 9,761 lower-body, and 9,568 dresses. All the images in MFVT are standardized to a resolution of $1024 \times 768$, and divided into training and test sets in a 7:3 ratio. Beyond MFVT, we further conduct experiments on two widely used open-sourced mask-based datasets: VTON-HD \cite{choi2021viton}, DressCode \cite{morelli2022dress}, and StreetVTON\cite{cui2025street}. VTON-HD is a high-resolution virtual try-on dataset that provides person–garment pairs with segmentation masks and emphasizes realistic garment transfer with fine-grained details. DressCode is a large-scale dataset focusing on fashion garments across upper-body, lower-body and dresses, offering diverse garment styles and person poses and widely adopted in prior virtual try-on research as a standard benchmark. StreetVTON support in-the-wild virtual try-on applications. In conclusion, we utilize MFVT for mask-free virtual try-on experiments, VTON-HD together with DressCode as representative mask-based benchmarks, and have demonstrated the scenario generalization of our method on StreetVTON.


\noindent\textbf{Implementation Details} 
We adopt Flux1.Fill as the backbone architecture for training. The IDT module is trained on two 80 GB NVIDIA A800 GPUs with a batch size of 16 for 15,000 steps. The OANO module is trained on a single 80 GB NVIDIA A800 GPU with a batch size of 64 for 150 epochs. 
We evaluate our model in two settings: mask-free and mask-based. In the mask-free setting, since only one open-source model is available, we evaluate exclusively on Any2AnyTryon \cite{guo2025any2anytryon}, which we retrain using MFVT to ensure a fair comparison. In the mask-based setting, we compare our method against eight state-of-the-art methods: StableVTON \cite{kim2024stableviton}, OOTDiffusion \cite{xu2025ootdiffusion}, CATVTON \cite{chong2024catvton}, TPD \cite{xu2025ootdiffusion}, IDM-VITON \cite{choi2024improving}, ITA-MDT \cite{hong2025ita}, OmniVTON\cite{yang2025omnivton} and FastFit\cite{chong2025fastfit}.
We report both quantitative and qualitative results. Specifically, Quantitative evaluation includes objective metrics such as LPIPS\cite{zhang2018unreasonable}, SSIM\cite{wang2004image}, FID\cite{heusel2017gans}, and KID\cite{kim2019u}, while qualitative evaluation is conducted through visual comparisons, highlighting perceptual realism, garment fidelity, and try-on consistency across diverse poses and identities.





\subsection{Quantitative Results}
Here, we present a comprehensive comparison of the quantitative results of our model against both mask-free and mask-based baselines. 

\noindent\textbf{Mask-free}
In the mask-free setting, we evaluate our framework against Any2Any-Tryon on the MFVT dataset, which includes upper-body, lower-body, and dresses categories. The evaluation employs four metrics: LPIPS, SSIM, FID, and KID. As reported in Tab.~\ref{tab:ab}, under the same number of inference steps, our method consistently surpasses Any2AnyTryon across all metrics. 
For example, at 6 inference steps, our method reduces FID by $2.906$, $3.136$ and $2.030$ for upper-body, lower-body and dresses, respectively. The results highlight the enhanced fidelity and perceptual similarity of the synthesized images, validating the effectiveness of our framework. Remarkably, our method achieves performance comparable to the 30-step results of Any2AnyTryon even with only 6 inference steps. 
For instance, in the dress category, Any2AnyTryon requires 30 steps to reach $0.829$ SSIM and $8.527$ FID, while our FDM-MFVT achieves $0.872$ SSIM and $8.431$ FID with only 6 steps. This highlights both the efficiency and practicality of our framework in mask-free scenarios.

\begin{table*}[]
\centering
\caption{Quantitative evaluation of our method against Any2AnyTryon on the MFVT dataset, including upper-body, lower-body, and dresses categories. Inf.Step refers
to the inference steps.}
\resizebox{\textwidth}{!}{
\begin{tabular}{c|c|cccc|cccc}
\toprule
\hline
\multirow{2}{*}{Method} & \multirow{2}{*}{Dataset}&\multicolumn{4}{c|}{Inf.Step=6} & \multicolumn{4}{c}{Inf.Step=30}\\ \cline{3-10}
&&LPIPS$\downarrow$ & SSIM$\uparrow$ & FID$\downarrow$ & KID$\downarrow$&LPIPS$\downarrow$ & SSIM$\uparrow$ & FID$\downarrow$ & KID$\downarrow$ \\
\hline
 & Upper-body &0.324&0.769&11.247&1.421 & 0.147 & 0.854 & 8.471 & 0.725 \\
 Any2AnyTryon  & Lower-body &0.475&0.774&11.452&1.213& 0.102 & 0.867 & 8.527 & 1.054 \\
  &  Dresses&0.348&0.771&10.461&1.411 & 0.164 & 0.829 & 8.527 & 0.947 \\
\hline
& Upper-body &0.087&0.887&8.341&0.725&0.085&0.893&8.341&0.712 \\
\textbf{FDM-MFVT (Ours)} & Lower-body &0.083&0.879&8.316&0.968 & 0.082 & 0.880 & 8.219 & 0.945 \\
& Dresses &0.102&0.872&8.431&0.901& 0.091 & 0.883 & 8.286 & 0.872 \\
\hline
\bottomrule
\end{tabular}}
\label{tab:ab}
\end{table*}

\noindent\textbf{Mask-based}
In the mask-based setting, we evaluate our method against eight state-of-the-art virtual try-on methods: StableVTON \cite{kim2024stableviton}, OOTDiffusion \cite{xu2025ootdiffusion}, CATVTON \cite{chong2024catvton}, TPD \cite{xu2025ootdiffusion}, IDM-VITON \cite{choi2024improving}, ITA-MDT \cite{hong2025ita}, OmniVTON\cite{yang2025omnivton} and FastFit\cite{chong2025fastfit}.
We report both quantitative and qualitative results. Specifically, Quantitative evaluation includes objective metrics such as LPIPS\cite{zhang2018unreasonable}, SSIM\cite{wang2004image}, FID\cite{heusel2017gans}, and KID\cite{kim2019u}. Experiments are conducted on the unpaired test sets of VTON-HD \cite{choi2021viton} and DressCode \cite{morelli2022dress} Upper-body, where the input garment differs from the target garment.The evaluation employs two metrics:FID, and KID. 
All models are tested under identical inference conditions, with comparisons made at both 6 and 30 inference steps to assess efficiency. As shown in Tab.~\ref{tab:ab1}, our method consistently outperforms all baselines across both FID and KID. Remarkably, despite not relying on explicit masks, our framework achieves superior performance compared to mask-based methods, demonstrating strong robustness and generalization ability.
For example, at 6 inference steps, our method surpasses the second-best baseline (ITA-MDT) by $2.367$ FID on VTON-HD \cite{choi2021viton} and by $9.206$ FID on DressCode \cite{morelli2022dress} Upper-body. Moreover, our 6-step outputs already match or exceed the 30-step results of competing methods. For instance, on DressCode Upper-body, our 6-step FID of $11.231$ is better than the 30-step FID of FastFit ($11.542$).
Together, the above results confirm that our method advances the state of the art in both effectiveness, efficiency and robustness for virtual try-on tasks, achieving higher fidelity and diversity while requiring fewer inference steps.

\begin{table*}[]
\centering
\caption{Quantitative comparison of our method and eight baselines on the unpaired test sets of VTON-HD \cite{choi2021viton} and DressCode \cite{morelli2022dress} Upper-body datasets. Inf.Step refers
to the inference steps. The best and second best results are in \textbf{bold} and \underline{underlined}. }
{
\resizebox{\textwidth}{!}{
\begin{tabular}{c|cc|cc|cc|cc}
\toprule
\hline \multirow{3}{*}{Method} & \multicolumn{4}{|c|}{VTON-HD} & \multicolumn{4}{|c}{DressCode Upper} \\ \cline{2-9}
& \multicolumn{2}{|c|}{Inf.Step=6}& \multicolumn{2}{|c|}{Inf.Step=30}& \multicolumn{2}{|c|}{Inf.Step=6}& \multicolumn{2}{|c}{Inf.Step=30}\\
& ~~FID $\downarrow$~~ & ~~KID $\downarrow$~~& ~~FID $\downarrow$~~ & ~~KID $\downarrow$~~& ~~FID $\downarrow$~~ & ~~KID $\downarrow$~~& ~~FID $\downarrow$~~ & ~~KID $\downarrow$~~  \\
\hline 
StableVTON & 13.427&3.427&9.717&1.450&22.847&6.547&13.666&3.857\\
OOTDDiffusion &16.362&5.871&12.268&3.214&37.106&7.839&29.383&4.302\\
CATVTON &13.949&3.946&10.263&2.370&23.185&3.147&12.720&1.910\\
TPD & 16.235&6.692&13.124&3.527&25.073&\underline{2.978}&12.957&1.874\\
IDM-VTON &  12.539&\underline{2.849}&9.261&\underline{1.253}&21.849&3.041&11.879&\underline{1.813} \\
ITA-MDT & \underline{11.054}&3.014&8.889&1.397&\underline{20.437}&3.256&11.928&1.871 \\
OmniVTON & 13.296&3.781&9.732&1.193&22.348&3.239&12.324&1.869\\
FastFit & 12.316&3.258&\underline{8.731}&1.264&21.316&3.124&\underline{11.542}&1.824\\
\textbf{FDM-MFVT (Ours) } &\textbf{8.687}&\textbf{1.201}&\textbf{8.659}&\textbf{1.010}&\textbf{11.231}&\textbf{1.401}&\textbf{10.963}&\textbf{1.394} \\
\hline
\bottomrule
\end{tabular}
}}
  \label{tab:ab1}
\end{table*}

\subsection{Qualitative Results}
\label{sec:qual}
\begin{figure}[!t]
	\centering
	\includegraphics[width=1\linewidth]{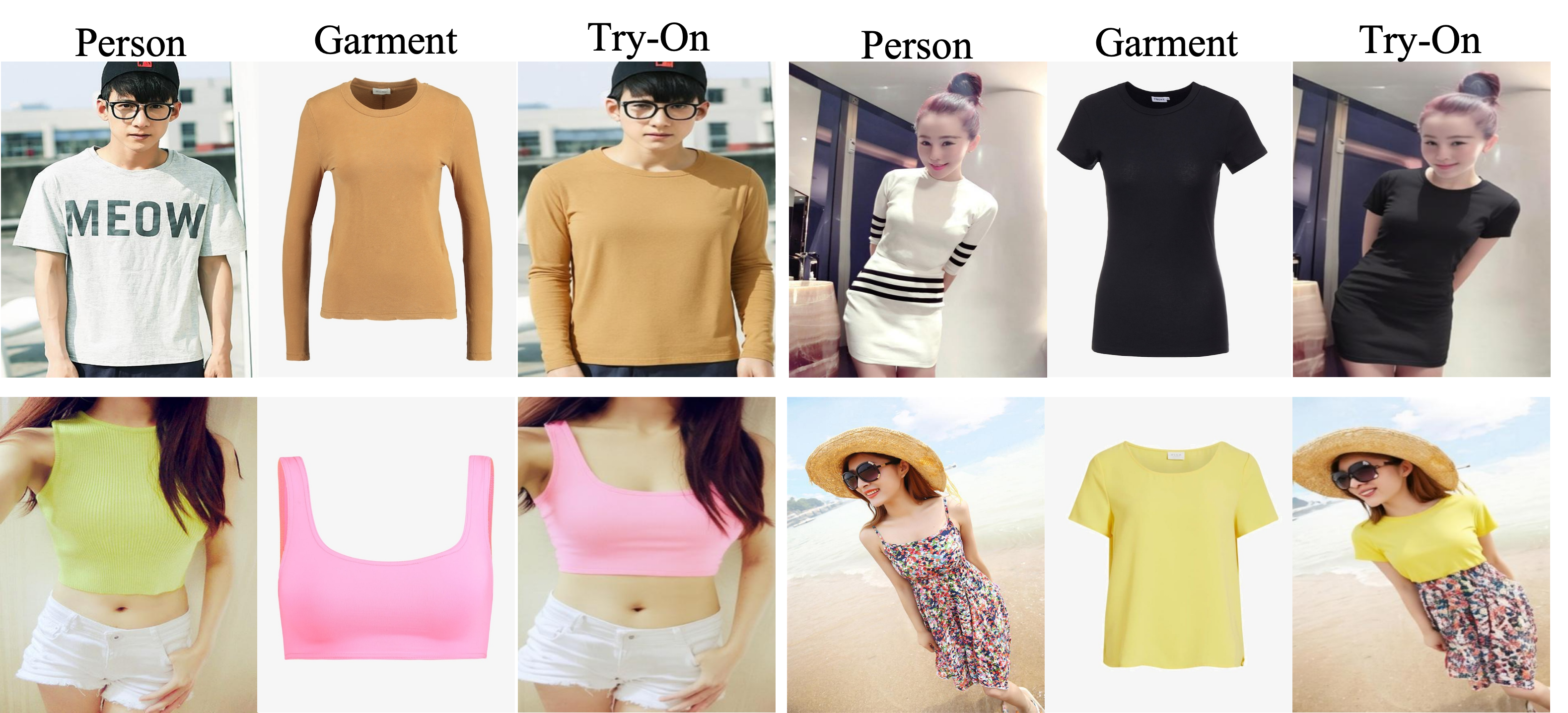}
	\caption{Qualitative results of our method on the StreetVTON dataset.
 }
	\label{fig:figure7}
\end{figure}

Here, we present a comprehensive comparison of qualitative results between our method and both mask-free and mask-based baselines, focusing on garment fidelity, texture realism, and overall visual coherence. Furthermore, we validated the scenario generalization ability of our method in the wild dataset as shown in Fig.~\ref{fig:figure7}. Fig.~\ref{fig:figure7} demonstrates our method’s ability to maintain garment fit and accessories using the StreetV-TON dataset. Our method achieves more accurate clothing rendering while retaining original image content for a more realistic result. Although our method was not trained in the wild dataset, Fig. 4 shows the high fidelity of the try-on image generated by our method in the wild dataset, demonstrating the generalization ability of our method.

\begin{figure*}[!t]
	\centering
	\includegraphics[width=1\linewidth]{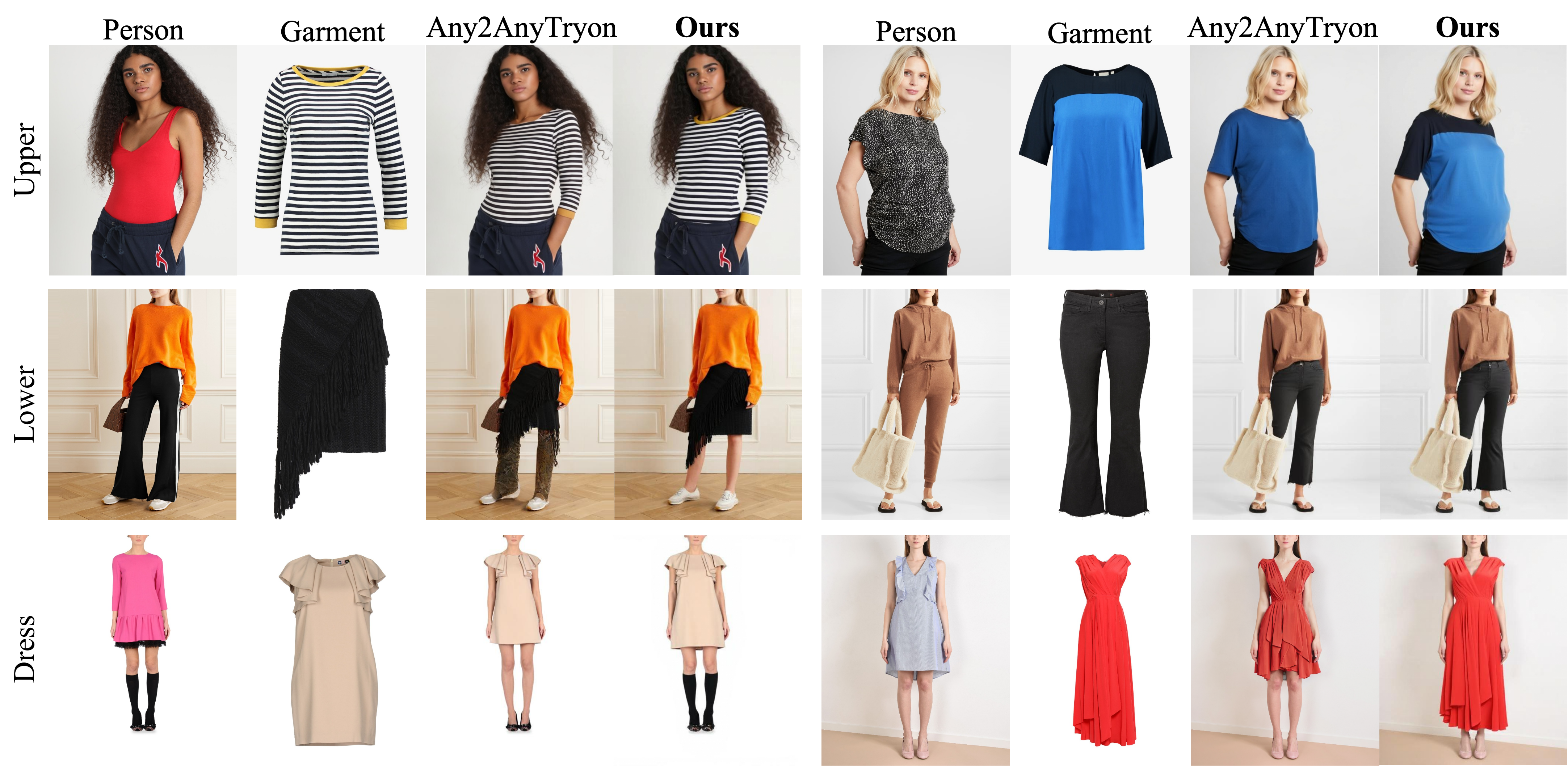}
	\caption{Qualitative comparison between our method and Any2AnyTryon on the MFVT dataset.}
	\label{fig:figure3}
\end{figure*}

\noindent\textbf{Mask-Free}
We conduct qualitative experiments on the MFVT dataset to compare our method with Any2AnyTryon across three garment categories: upper-body, lower-body, and dresses. As shown in Fig.~\ref{fig:figure3}, each row corresponds to one category, with two representative cases provided per row. The first row showcases upper-body garments such as tank tops and long-sleeve shirts, the second row presents lower-body garments including pants and skirts, and the third row displays dresses with varied styles and textures.
Our method consistently produces more realistic and coherent try-on results, accurately preserving garment-specific attributes such as texture, length, and material appearance. For instance, 
in the second row, our output preserves the color contrast and structural details of the orange top, whereas Any2AnyTryon introduces undesired black-pants artifacts originating from the initial model’s garment.
The enhanced garment-body alignment, texture realism, and structural consistency achieved by our method highlight its effectiveness in mask-free virtual try-on scenarios, especially under challenging conditions involving complex patterns and full-body poses.

\begin{figure*}[!t]
	\centering
	\includegraphics[width=1\linewidth]{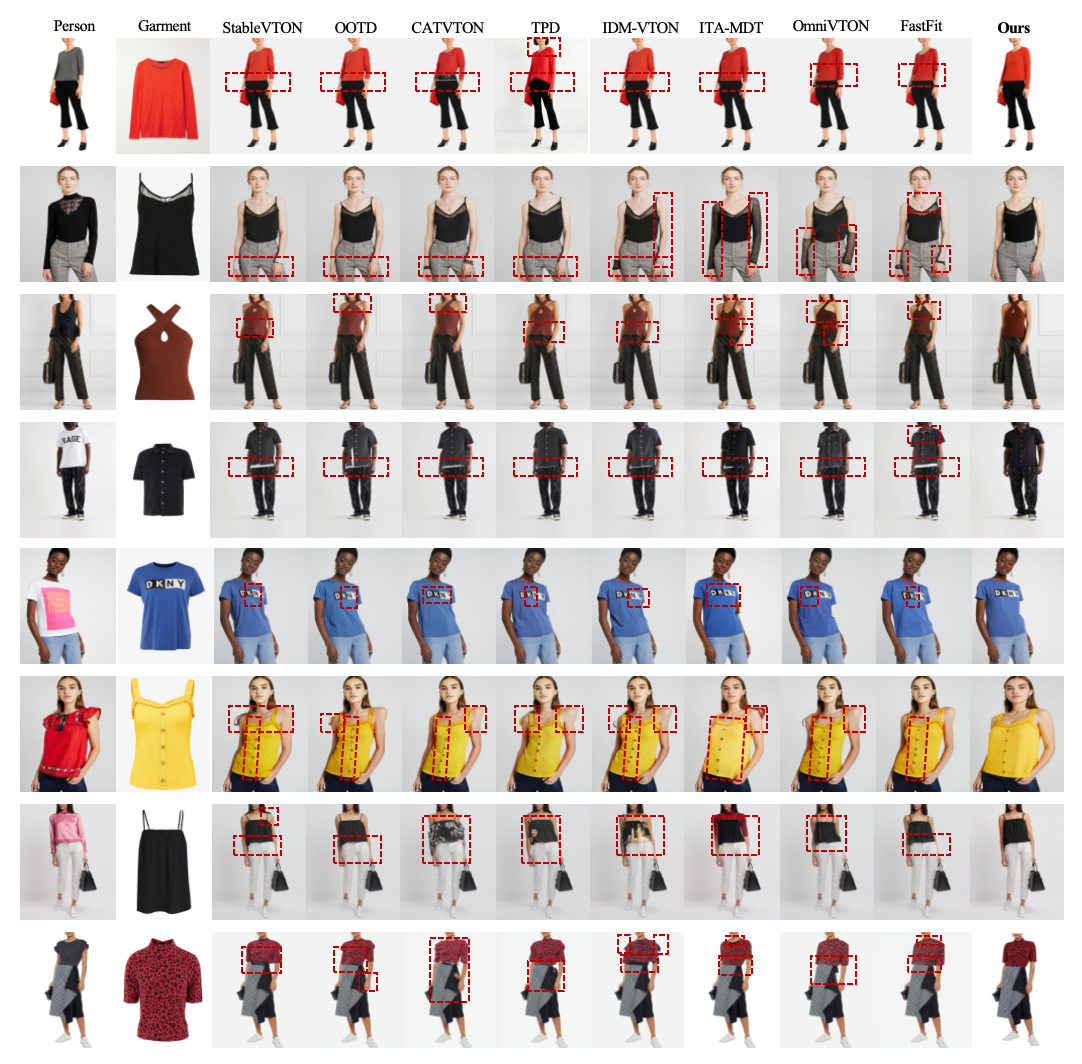}
	\caption{Qualitative comparison between our method and eihgt baselines on the unpaired test sets of VTON-HD \cite{choi2021viton} and DressCode \cite{morelli2022dress} Upper-body datasets.
 }
	\label{fig:figure4}
\end{figure*}

\noindent\textbf{Mask-based} 
To assess performance in mask-based scenarios, we conduct comparisons between our method and six leading baselines: StableVTON \cite{kim2024stableviton}, OOTDiffusion \cite{xu2025ootdiffusion}, CATVTON \cite{chong2024catvton}, TPD \cite{xu2025ootdiffusion}, IDM-VITON \cite{choi2024improving}, ITA-MDT \cite{hong2025ita}, OmniVTON\cite{yang2025omnivton} and FastFit\cite{chong2025fastfit}. All evaluations are performed on the unpaired test sets of the VTON-HD \cite{choi2021viton} and DressCode \cite{morelli2022dress} datasets. As illustrated in Fig.~\ref{fig:figure4}, our method consistently generates more photo-realistic try-on results with sharper textures and cleaner garment boundaries. For instance, in the second row, the black lace camisole preserves its intricate detailing and layered structure in our output, whereas the baseline results exhibit distortions around the wrist, partially retaining elements of the original garment and even hallucinating additional hair, largely due to poor mask quality. Overall, these comparisons highlight the robustness of our framework in mask-based scenarios, achieving superior visual fidelity and garment consistency over existing baselines.

\begin{table}[]
\centering
\caption{Quantitative results of FDM-MFVT ablation.}
{
\begin{tabular}{l|ccccc}
\toprule
\hline 
Method & LPIPS $\downarrow$ & SSIM $\uparrow$  &FID $\downarrow$ & KID $\downarrow$\\
\hline 
w/o OANO & 0.121&0.864&8.417&1.019 \\
w/o IDT & 0.135&0.859&8.378&1.142\\
\textbf{FDM-MFVT} & \textbf{0.091}&\textbf{0.879}&\textbf{8.363}&\textbf{0.865}\\
\hline
\bottomrule
\end{tabular}
}
  \label{tab:ablation}
\end{table}

\subsection{Ablation Study}
\begin{figure}[!t]
	\centering
	\includegraphics[width=1\linewidth]{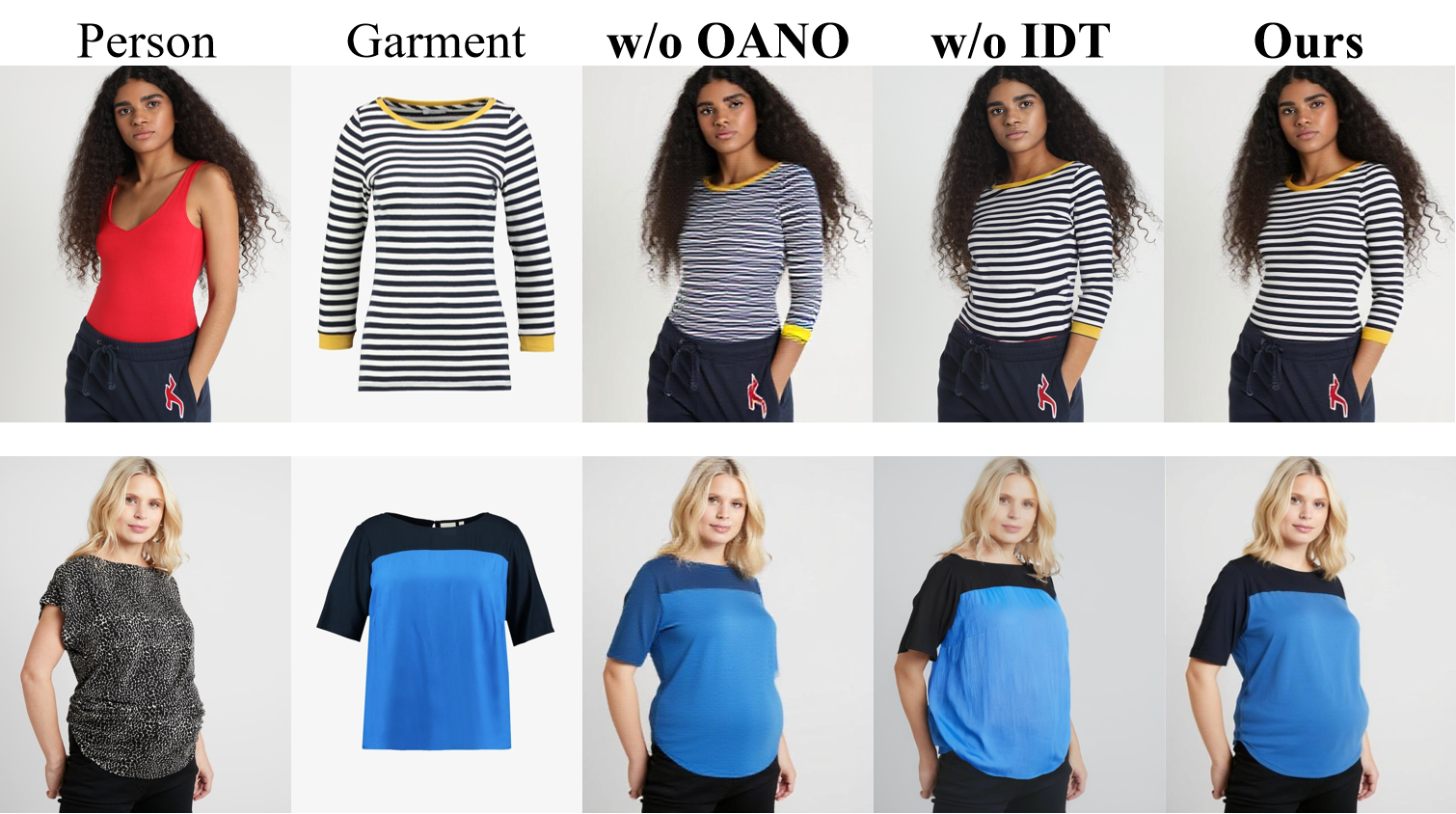}
	\caption{Qualitative results of FDM-MFVT ablation.
 }
	\label{fig:figure5}
\end{figure}

\noindent\textbf{Ablation Study}
To evaluate the contribution of each component in our FDM-MFVT framework, we conduct an ablation study by selectively removing the OANO module and the IDT module. As shown in Table~\ref{tab:ablation}, removing OANO leads to a noticeable degradation in perceptual quality, with LPIPS increasing from $0.091$ to $0.121$ and SSIM dropping from $0.879$ to $0.864$. This confirms the importance of noise refinement in enhancing structural consistency and visual realism. Similarly, excluding IDT module results in further performance decline, with $0.135$ and $0.859$ decline in  LPIPS and SSIM, indicating that instruction-guided alignment plays a crucial role in generating coherent garment-body compositions.  
The full model achieves the best results across all metrics, including the lowest FID of $8.363$ and KID of $0.865$, demonstrating the effectiveness of integrating both modules. Qualitative comparisons in Fig.~\ref{fig:figure5} further support these results. Models without OANO or IDT module exhibit noticeable artifacts, misaligned garments, and reduced texture fidelity, whereas our complete framework produces visually consistent and realistic try-on results.

\noindent\textbf{Efficiency Analysis}
The OANO module significantly reduced the number of inference steps and obtained higher fidelity virtual try-on images in just 0.28s. In addition, the OANO module has a small number of parameters and is easy to train. These factors highlight the lightweight, efficient, and broad application potential of OANO module.

\section{Conclusion}
In this work, we presented FDM-MFVT, a novel few-step diffusion framework for mask-free image-based virtual try-on. By integrating an OANO module and an IDT module, our method significantly reduced sampling cost while maintaining high-quality synthesis. The OANO module enabled efficient convergence by aligning the initial noise with the input image, achieving comparable results to 30-step baselines with only 6 inference steps. The IDT module further enhanced flexibility by leveraging garment-person inputs without relying on segmentation masks or auxiliary networks.
To support mask-free training and evaluation, we introduced MFVT, a large-scale paired dataset comprising $30,000$ high-resolution garment-person pairs across diverse categories. Extensive experiments on MFVT, VTON-HD \cite{choi2021viton}, DressCode \cite{morelli2022dress} and StreetVTON\cite{cui2025street}demonstrated that FDM-MFVT consistently outperformed both mask-free and mask-based baselines in terms of fidelity, diversity, and efficiency. Ablation studies confirmed the effectiveness of each module, and qualitative comparisons highlighted superior texture preservation and garment-body alignment.
Together, our contributions advanced the state of the art in virtual try-on by addressing key limitations of existing diffusion-based methods, offering a scalable, robust, and real-time solution for mask-free try-on synthesis.

\section*{Acknowledgements}
This work was supported by the National Natural Science Foundation of China under Grants 62401027 and 62231002.

%
%
\bibliographystyle{splncs04}
\bibliography{main}

%
%
\end{document}